\documentclass{article}

\PassOptionsToPackage{numbers, compress}{natbib}
 \usepackage[preprint]{neurips_2026}

\usepackage[utf8]{inputenc} 
\usepackage[T1]{fontenc}    
\usepackage{hyperref}       
\usepackage{url}            
\usepackage{booktabs}       
\usepackage{amsfonts}       
\usepackage{nicefrac}       
\usepackage{microtype}      
\usepackage{xcolor}         

\usepackage{soul}

\title{GEM: Gaussian Evolution Model for Occupancy Forecasting and Motion Planning}

\author{%
  Cheng Chen \\
  Purdue University\\
  \texttt{chen4384@purdue.edu} \\
  \And
  Hao Huang \\
  New York University Abu Dhabi \\
  \texttt{hh1811@nyu.edu} \\
  \And
  Saurabh Bagchi \\
  Purdue University \\
  \texttt{sbagchi@purdue.edu} \\
}

\input{headers}
\begin{document}

\maketitle

\begin{abstract}
Future 3D semantic occupancy forecasting and motion planning are central to autonomous driving, as they require models to reason about how surrounding scenes evolve and how the ego vehicle should act. Existing occupancy world models commonly discretize scenes into latent embeddings, volumetric features, or quantized tokens, and forecast future states through fixed-step autoregressive generation. This limits temporal flexibility, obscures scene evolution, accumulates errors over long horizons, and poorly matches the continuous-time dynamics of real driving scenes. We propose \textbf{\name}, a \textbf{G}aussian \textbf{E}volution \textbf{M}odel for non-autoregressive occupancy world modeling, where driving scenes are represented as explicit continuous 4D Gaussian primitives with learned dynamics. Instead of rolling out future occupancy states step by step, \name directly queries the Gaussian world representation at arbitrary timestamps and splats the corresponding conditional 3D Gaussians into semantic occupancy volumes. This enables efficient forecasting over the full horizon while retaining a compact and interpretable scene representation. By decoupling spatial geometry, temporal support, and primitive motion, \name makes the predicted world easier to inspect, as each primitive's evolution can be followed continuously over time. The same representation also supports motion planning by predicting future ego trajectories from the learned Gaussian world. Extensive experiments show that \name achieves state-of-the-art future semantic occupancy forecasting and strong motion planning performance, while providing flexible temporal querying.

\end{abstract}
\section{Introduction}
\label{sec:intro}

\begin{figure}[tb]
  \centering
  \includegraphics[width=.99\columnwidth]{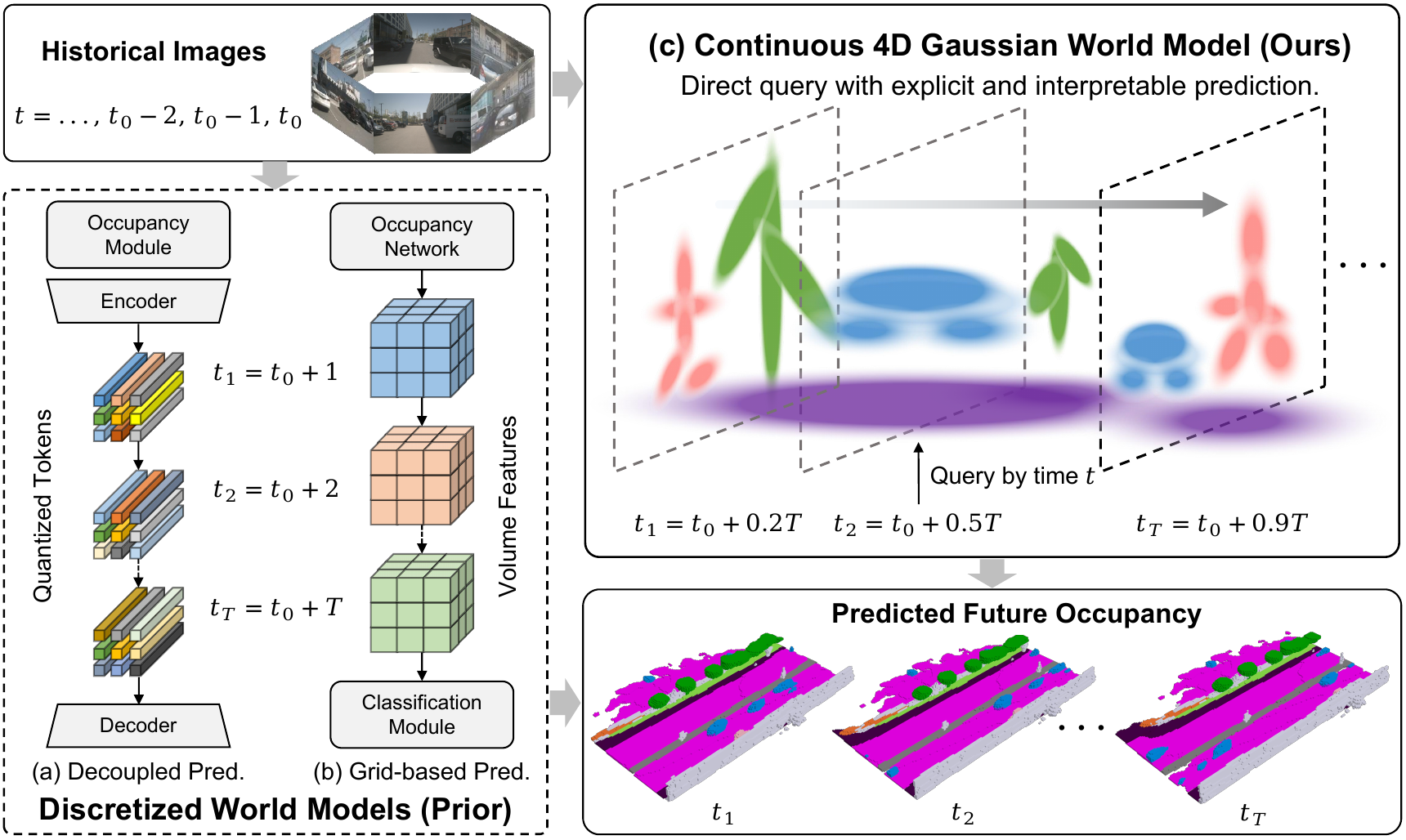}
  \caption{Comparison between \textit{discretized} occupancy world models and our \textit{continuous} 4D Gaussian world model. Existing methods typically forecast future occupancy through (a) discrete tokens or (b) volumetric features at fixed timestamps, often requiring sequential rollout. In contrast, our \textbf{\name} shown in (c) represents the scene as evolving 4D semantic Gaussian primitives that can be directly queried at arbitrary timestamps and splatted into future semantic occupancy volumes, enabling explicit, interpretable, and non-autoregressive scene forecasting.
  }
  \label{fig:teaser}
\end{figure}

Autonomous driving requires models to perceive the surrounding scene, forecast its future evolution, and plan safe ego trajectories. Vision-centric end-to-end methods~\cite{jiang2023vad,wu2022tcp} directly predict future trajectories from monocular or multi-view camera images, but their black-box nature limits interpretability and complicates safety analysis. To introduce more structured reasoning, intermediate representations such as Bird's-Eye View (BEV) have been widely adopted~\cite{uniad,stp3}. However, BEV often compresses the vertical dimension, making it less suitable for fine-grained 3D scene understanding. Semantic occupancy addresses this limitation by representing 3D space with dense geometric and semantic labels, and has shown strong potential for planning. Building on this representation, \textit{4D occupancy forecasting}~\cite{zheng2024occworld} further predicts future semantic occupancy from historical observations, providing a predictive world model for autonomous driving.

Existing vision-centric occupancy world models~\cite{zheng2024occworld,occllm,yan2025renderworld,li2025semisupervised,dang2026sparseworld} typically encode the observed 3D scene into latent states, such as discrete tokens (Figure~\ref{fig:teaser}-(a)), dense volumetric features (Figure~\ref{fig:teaser}-(b)), or sparse point queries, and then forecast future states, often sequentially in an autoregressive manner, before decoding them into occupancy. \textit{Despite promising progress, existing occupancy world models still face three formidable limitations.} \textit{First}, these works mainly rely on multi-stage pipelines with separate perception, embedding, and forecasting modules, making end-to-end optimization more complex. \textit{Second}, their encode--predict--decode design models future dynamics in latent space rather than through an explicit and interpretable scene representation. \textit{Third}, autoregressive forecasting restricts prediction to predefined time steps and requires all intermediate future states to be generated sequentially before reaching a target horizon; thus, farther-time queries require multiple dependent rollouts, increasing inference latency and computation. These limitations reduce both pipeline simplicity and temporal flexibility for downstream tasks.

To address these limitations, we introduce \textbf{\name}, a continuous, non-autoregressive occupancy world model built on evolving 4D semantic Gaussian primitives. Instead of predicting future states through discrete representations and step-wise rollout, \name models scene evolution directly in continuous spatiotemporal space. As shown in Figure~\ref{fig:teaser}-(c), each primitive is grounded in space and time and carries explicit geometric, semantic, temporal, and motion attributes. Future occupancy can therefore be queried at \textit{arbitrary} timestamps by slicing and splatting the corresponding time-conditioned Gaussians, without sequential decoding over fixed time steps. This formulation simplifies forecasting and provides a flexible representation for future scene occupancy prediction. Specifically, \name uses a forecasting-oriented 4D primitive representation. Rather than learning a generic coupled spatiotemporal covariance, each primitive explicitly separates spatial geometry, temporal support, and motion. This structured parameterization is easier to optimize and inspect for dynamic occupancy forecasting. Furthermore, to represent future scenes in the moving ego coordinate frame, \name injects ego-vehicle state context into both Gaussian evolution and trajectory prediction. This allows each Gaussian primitive's velocity to be decomposed into ego-induced scene motion and object-level (\eg, pedestrains, bicycles, trucks, cars, \etc) dynamic motion within a unified Gaussian world representation.

By combining these designs, \name achieves state-of-the-art vision-centric 4D occupancy forecasting, outperforming prior occupancy world models, such as OccWorld~\cite{zheng2024occworld}, PreWorld~\cite{li2025semisupervised}, and SparseWorld~\cite{dang2026sparseworld}. It also delivers comparable or better motion planning performance, with competitive trajectory accuracy and lower collision rates, while retaining an explicit and continuously queryable Gaussian world representation. We summarize our contributions are as follows:

\begin{itemize}[leftmargin=*]
    \item We introduce \textbf{\name}, a vision-centric, non-autoregressive occupancy world model that represents dynamic driving scenes as evolving 4D semantic Gaussian primitives.
    
    \item We propose a structured continuous-time Gaussian formulation that decouples spatial geometry, temporal support, semantics, opacity, and motion, enabling arbitrary-time occupancy querying without sequential rollout over fixed future steps.

    \item We design an ego-conditioned Gaussian world framework that decomposes Gaussian primitive motion into ego-induced and object-level components, and supports ego vehicle motion planning from the same unified Gaussian representation.
    
    \item We validate \name on the Occ3D-nuScenes benchmark~\cite{tian2023occ3d}, where it achieves state-of-the-art 4D occupancy forecasting and strong motion planning performance compared with existing vision-centric occupancy world models.
\end{itemize}
\section{Methodology}
\label{sec:method}

\subsection{Preliminary}
For a vehicle at timestamp $t$, vision-centric \textit{3D occupancy prediction} takes $N$ synchronized camera views $\mathbf{S}_t=\{I_t^{(1)}, I_t^{(2)}, \ldots, I_t^{(N)}\}$ as input and estimates the semantic occupancy of the current 3D scene. The output is a voxelized semantic occupancy volume $\mathbf{Y}_t \in [0,1]^{X \times Y \times Z \times C}$,
where $(X,Y,Z)$ denotes the spatial resolution of the discretized 3D volume, and $C$ is the number of semantic categories. Each voxel therefore stores a categorical occupancy distribution over the semantic classes. In contrast, vision-centric \textit{4D occupancy forecasting} extends 3D occupancy prediction from current scene reconstruction to future scene evolution. Given a temporal sequence of past multi-view observations $\mathbf{S}_{t-K:t} \triangleq \{\mathbf{S}_{t-K}, \ldots, \mathbf{S}_{t-1}, \mathbf{S}_t\}$, the model forecasts a sequence of future 3D semantic occupancy volumes over the next $T$ timestamps $\mathbf{Y}_{t+1:t+T} \triangleq \{\mathbf{Y}_{t+1}, \mathbf{Y}_{t+2}, \ldots, \mathbf{Y}_{t+T}\}$. Here, each $\mathbf{Y}_{t+i}$ denotes the predicted semantic occupancy volume at future timestamp $t+i$.

For \textit{motion planning}, the end-to-end autonomous driving model predicts the future ego trajectory as a sequence of planned waypoints $\boldsymbol{\tau}_{t+1:t+T} \triangleq \{\boldsymbol{\tau}_{t+1}, \boldsymbol{\tau}_{t+2}, \ldots, \boldsymbol{\tau}_{t+T}\}$. Here, each $\boldsymbol{\tau}_{t+i}$ denotes the predicted future position of the ego vehicle at timestamp $t+i$, typically represented in the ego-centric ground-plane coordinate system. Together, these waypoints define the planned trajectory over the forecasting horizon.

A practical occupancy-based driving world model $\mathcal{W}$ should jointly reason about future scene evolution and ego-vehicle planning. To this end, we formulate $\mathcal{W}$ as a conditional predictive model that takes historical multi-view observations and historical ego-state context as input. The ego-state context may include velocity, acceleration, and past ego positions. Let $
\mathbf{E}_{t-K:t} \triangleq \{\mathbf{e}_{t-K}, \ldots, \mathbf{e}_{t-1}, \mathbf{e}_t\}$
denote the historical ego-state sequence. The world model is then defined as:
\begin{equation}
\mathcal{W}:\; 
(\mathbf{S}_{t-K:t}, \mathbf{E}_{t-K:t})
\;\mapsto\;
\big(\mathbf{Y}_{t+1:t+T}, \boldsymbol{\tau}_{t+1:t+T}\big)\enspace.
\label{eq:world_model_mapping}
\end{equation}
Jointly predicting $\mathbf{Y}_{t+1:t+T}$ and $\boldsymbol{\tau}_{t+1:t+T}$ encourages consistency between the forecasted scene dynamics and the ego vehicle's planned motion.

\subsection{Continuous Decoupled 4D Gaussian World Model}
\label{subsec:continuous_4d_world_model}

To obtain a continuous and interpretable representation of future scene evolution, we model the driving world as a set of evolving Gaussian primitives in joint space-time, which is essentially different from autoregressive occupancy world models, which typically predict future voxel grids~\cite{li2025semisupervised}, BEV features~\cite{driveoccworld}, discrete tokens~\cite{zheng2024occworld,yan2025renderworld,liao2025i2,occllm,wei2024occllama}, or latent queries~\cite{dang2026sparseworld} step by step at a fixed set of future timestamps. Such autoregressive designs must generate intermediate states before reaching a target horizon, which restricts temporal flexibility and may accumulate prediction errors over long rollouts. Recent 4D Gaussian representations have shown that continuous space-time primitives provide a natural representation for dynamic scenes in novel view synthesis~\cite{yang2023gs4d,4drotor,asiimwe20254d}. Inspired by these works, we represent the entire forecasting horizon with a unified continuous \textit{4D Gaussian world model} for semantic occupancy prediction, where time-conditioned Gaussian primitives are sliced and splatted into occupancy volumes for arbitrary-time future prediction. As each primitive carries explicit spatial geometry, temporal support, semantic attributes, and motion dynamics, the future scene can be directly queried at any timestamp without sequential rollout.

Analogous to 3D Gaussian primitives~\cite{3dgs}, each 4D primitive in dynamic scenes is defined over a 4D space-time coordinate $\mathbf{u}=[\mathbf{x}^{\top}, t]^{\top}\in\mathbb{R}^{4}$, with a 4D mean $\boldsymbol{\mu}_{\ssFourD}=[\boldsymbol{\mu}_s^{\top}, \mu_t]^{\top}$ and covariance matrix $\boldsymbol{\Sigma}_{\ssFourD} \in \mathbb{R}^{4\times 4}$. The corresponding Gaussian density can be written as:
\begin{equation}
\mathcal{G}(\mathbf{u}) =
\exp\left(
-\frac{1}{2}
(\mathbf{u}-\boldsymbol{\mu}_{\ssFourD})^\top
\boldsymbol{\Sigma}_{\ssFourD}^{-1}
(\mathbf{u}-\boldsymbol{\mu}_{\ssFourD})
\right)\enspace.
\end{equation}
The covariance can be parameterized through a 4D rotation-scale decomposition:
\begin{equation}
\boldsymbol{\Sigma}_{\ssFourD}
=
\mathbf{R}_{\ssFourD}\mathbf{S}_{\ssFourD}\mathbf{S}_{\ssFourD}^{\top}\mathbf{R}_{\ssFourD}^{\top}\enspace,
\label{eq:cov}
\end{equation}
\begin{wrapfigure}{r}{0.5\columnwidth}
    \centering
    \includegraphics[width=0.48\columnwidth]{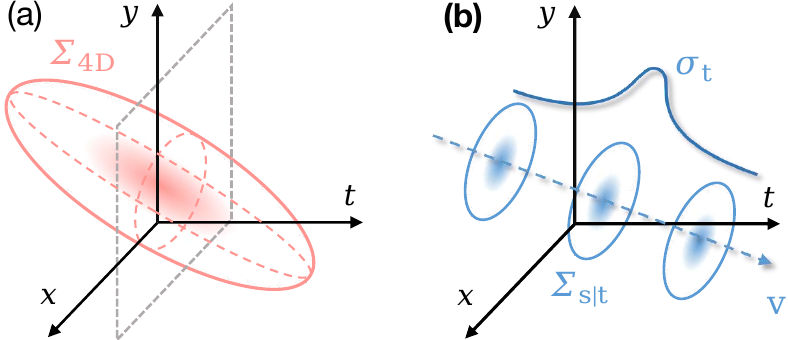}
    \caption{Unlike a full 4D covariance shown in (a) that couples space and time, our structured formulation shown in (b) decouples spatial covariance, temporal support, and motion for interpretable continuous occupancy forecasting. For visualization, we only show the $x$-$y$ plane and omit the $z$-axis.}
    \label{fig:reformulation}
\end{wrapfigure}
where $\mathbf{S}_{\ssFourD}=\operatorname{diag}(s_x,s_y,s_z,s_t)$ contains the spatial and temporal scales, and $\mathbf{R}_{\ssFourD}$ denotes a 4D rotation matrix, which can be parameterized by either a pair of 4D isotropic rotations~\cite{yang2023gs4d} or 4D rotors~\cite{4drotor}. Although this full 4D covariance formulation is expressive, it allows arbitrary coupling between spatial and temporal axes, as shown in Figure~\ref{fig:reformulation}-(a). However, for occupancy forecasting in driving scenes, such unconstrained space-time mixing is difficult to interpret and imposes unnecessary optimization burden. This motivates our structured conditional formulation, which decouples spatial geometry, temporal support, and primitive motion instead of learning an unrestricted 4D covariance as in Equation~\eqref{eq:cov}.

Formally, rather than parameterizing each primitive with a 4D scale matrix $\mathbf{S}_{\ssFourD}$ and a 4D rotation matrix $\mathbf{R}_{\ssFourD}$ as in prior 4D Gaussian representations~\cite{yang2023gs4d,4drotor}, we decompose the 4D covariance into spatial and temporal components. Specifically, we write 
$
\mathbf{\Sigma}_{\ssFourD}
=
\begin{bmatrix}
\mathbf{\Sigma}_{ss} & \mathbf{\Sigma}_{st}\\
\mathbf{\Sigma}_{ts} & \sigma_t^2
\end{bmatrix}
$, where $\mathbf{\Sigma}_{ss} \in \mathbb{R}^{3\times 3}$ describes the spatial covariance, $\sigma_t^2 \in \mathbb{R}$ denotes the temporal variance, and $\mathbf{\Sigma}_{st} \in \mathbb{R}^{3\times 1}$ captures the space-time correlation. Under this formulation, a 4D Gaussian primitive can be sliced at any queried timestamp by considering the conditional distribution of space given time. Applying the standard Gaussian conditioning rule yields:
\begin{equation}
\mathbf{x}\mid t
\sim
\mathcal{N}\!\left(
\boldsymbol{\mu}_s + \mathbf{\Sigma}_{st}\sigma_t^{-2}(t-\mu_t),\;
\mathbf{\Sigma}_{ss} - \mathbf{\Sigma}_{st}\sigma_t^{-2}\mathbf{\Sigma}_{ts}
\right)\enspace.
\label{eq:gaussian_conditioning}
\end{equation}
An important property of Equation~\eqref{eq:gaussian_conditioning} is that the conditional covariance is independent of the queried timestamp $t$; only the conditional mean varies with time. This property motivates our structured parameterization: instead of learning an unrestricted 4D covariance using Equation~\eqref{eq:cov} and indirectly recovering the sliced 3D Gaussian using Equation~\eqref{eq:gaussian_conditioning}, we directly predict the time-invariant conditional spatial covariance together with an explicit temporal motion term. This yields a more interpretable and easier-to-optimize representation for continuous occupancy forecasting.

Concretely, as shown in Figure~\ref{fig:reformulation}-(b), we avoid recovering the sliced 3D Gaussian from an unrestricted 4D covariance. Instead, we directly learn the spatial conditional covariance $\mathbf{\Sigma}_{s\mid t}$, the temporal scale $\sigma_t$, and the motion coefficient $\mathbf{v}$. The time-conditioned primitive in Equation~\eqref{eq:gaussian_conditioning} can be written as:
\begin{equation}
\mathbf{x}\mid t
\sim
\mathcal{N}\!\left(
\boldsymbol{\mu}_s + \mathbf{v}(t-\mu_t),\;
\mathbf{\Sigma}_{s\mid t}
\right)\enspace.
\label{eq:conditional_model}
\end{equation}
This parameterization gives each quantity a clear physical meaning. The spatial mean $\boldsymbol{\mu}_s$ denotes the primitive center at the reference time $\mu_t$. The vector $\mathbf{v}$ describes how the primitive center moves as the query time changes. The conditional covariance $\mathbf{\Sigma}_{s\mid t}$ specifies the spatial extent and orientation of the sliced 3D Gaussian, and is shared across queried timestamps. The temporal variance $\sigma_t^2$ determines the temporal support of the primitive, \ie, how strongly it contributes when the queried timestamp $t$ moves away from $\mu_t$ through a Gaussian temporal weight. As a result, each primitive forms a continuous, localized, and motion-aware scene element for arbitrary-time occupancy forecasting.

Although Equation~\eqref{eq:conditional_model} is expressed as a time-conditioned 3D Gaussian, it is still consistent with a valid joint 4D Gaussian. Specifically, given the learned conditional spatial covariance $\mathbf{\Sigma}_{s\mid t}$, motion coefficient $\mathbf{v}$, and temporal variance $\sigma_t^2$, we construct the corresponding joint space-time covariance:
\begin{equation}
\mathbf{\Sigma}_{\ssFourD}
=
\begin{bmatrix}
\mathbf{\Sigma}_{s\mid t} + \sigma_t^2\mathbf{v}\mathbf{v}^{\top} 
& 
\sigma_t^2\mathbf{v}\\
\sigma_t^2\mathbf{v}^{\top} 
& 
\sigma_t^2
\end{bmatrix}\enspace.
\label{eq:structured_joint_covariance}
\end{equation}
Under this construction, the space-time cross-covariance is $\mathbf{\Sigma}_{st}=\sigma_t^2\mathbf{v}$, while the spatial covariance of the joint 4D Gaussian is $\mathbf{\Sigma}_{ss}=\mathbf{\Sigma}_{s\mid t}+\sigma_t^2\mathbf{v}\mathbf{v}^{\top}$. Substituting these terms into the Gaussian conditioning formula in Equation~\eqref{eq:gaussian_conditioning} recovers exactly the conditional model in Equation~\eqref{eq:conditional_model}. Therefore, our formulation is not a heuristic motion parameterization; instead, it defines a structured family of valid 4D Gaussians in which the conditional mean evolves linearly with time, while the conditional spatial covariance remains time-invariant.

We further constrain the motion term according to the geometry of driving scenes. While $\mathbf{v}\in\mathbb{R}^3$ could model motion along all spatial axes, scene dynamics in autonomous driving are dominated by ground-plane motion, and vertical displacement is usually limited. Allowing unrestricted $z$-axis motion therefore introduces an unnecessary degree of freedom that reduce training stability and make the learned primitive motion less interpretable. We therefore adopt a planar velocity parameterization:
\begin{equation}
\mathbf{v} =
\begin{bmatrix}
v_x & v_y & 0
\end{bmatrix}^{\top}\enspace,
\label{eq:planar_velocity}
\end{equation}
which improves interpretability by making temporal evolution shift each primitive center only in the ground plane. The vertical spatial extent is still captured by the conditional covariance $\mathbf{\Sigma}_{s\mid t}$, and the formulation can be extended to full 3D velocity when necessary.

For the conditional spatial covariance, we follow GaussianFormer~\cite{huang2024gaussianformer} and use a 3D rotation-scale decomposition:
\begin{equation}
\mathbf{\Sigma}_{s\mid t}
=
\mathbf{R}_{\ssThreeD}
\mathbf{S}_{\ssThreeD}
\mathbf{S}_{\ssThreeD}^{\top}
\mathbf{R}_{\ssThreeD}^{\top}\enspace,
\label{eq:conditional_covariance_param}
\end{equation}
where $\mathbf{S}_{\ssThreeD}=\operatorname{diag}(s_x,s_y,s_z)$ defines the Gaussian scales, and $\mathbf{R}_{\ssThreeD}\in SO(3)$ is obtained from a quaternion parameterization. Unlike full 4D rotation-based formulations that mix spatial and temporal axes, this decomposition keeps the primitive geometry purely spatial: $\mathbf{\Sigma}_{s\mid t}$ determines the spatial extent and orientation of the sliced 3D Gaussian at any queried timestamp, while $\mathbf{v}$ controls how its center moves over time and $\sigma_t$ controls its temporal support.

At inference time, we query the continuous 4D Gaussian world by slicing the same primitives at any timestamp $t$ within the forecasting horizon. Each primitive yields a conditional 3D Gaussian with mean
$
\boldsymbol{\mu}_{s\mid t}
=
\boldsymbol{\mu}_s + \mathbf{v}(t-\mu_t)
$
and covariance $\mathbf{\Sigma}_{s\mid t}$. Its temporal support is measured by:
\begin{equation}
w(t)
=
\exp\!\left(
-\frac{(t-\mu_t)^2}{2\sigma_t^2}
\right)\enspace,
\label{eq:temporal_weight}
\end{equation}
which modulates the opacity $o$ to form the effective splatting weight $\tilde{o}(t)=o\cdot w(t)$. The resulting time-conditioned Gaussians are splatted with $\tilde{o}(t)$ and semantic logits $\mathbf{c}$ to produce the semantic occupancy prediction at timestamp $t$. Since $\boldsymbol{\mu}_{s\mid t}$, $\mathbf{\Sigma}_{s\mid t}$ and $w(t)$ are defined over continuous time, our \name can be evaluated at arbitrary temporal resolutions without changing the representation or sequentially rolling out states. This essentially differs from autoregressive occupancy world models~\cite{zheng2024occworld,wei2024occllama,li2025semisupervised,yan2025renderworld,liao2025i2,occllm,driveoccworld,dang2026sparseworld}, which typically predict future states at fixed time steps and must generate preceding states before reaching a target horizon.

\noindent \textbf{Remark.} This representation is well suited for semantic occupancy forecasting. It supports continuous-time querying, exposes interpretable primitive attributes including spatial geometry, temporal support, semantics, and motion, and models future evolution through explicit primitive motion rather than opaque latent transitions. For efficiency, \name uses a single shared set of 4D Gaussians to describe the full forecasting horizon, avoiding separate latent states for individual future steps while remaining compatible with voxel-space supervision and Gaussian-to-voxel splatting.

\begin{figure}
    \centering
    \includegraphics[width=\columnwidth]{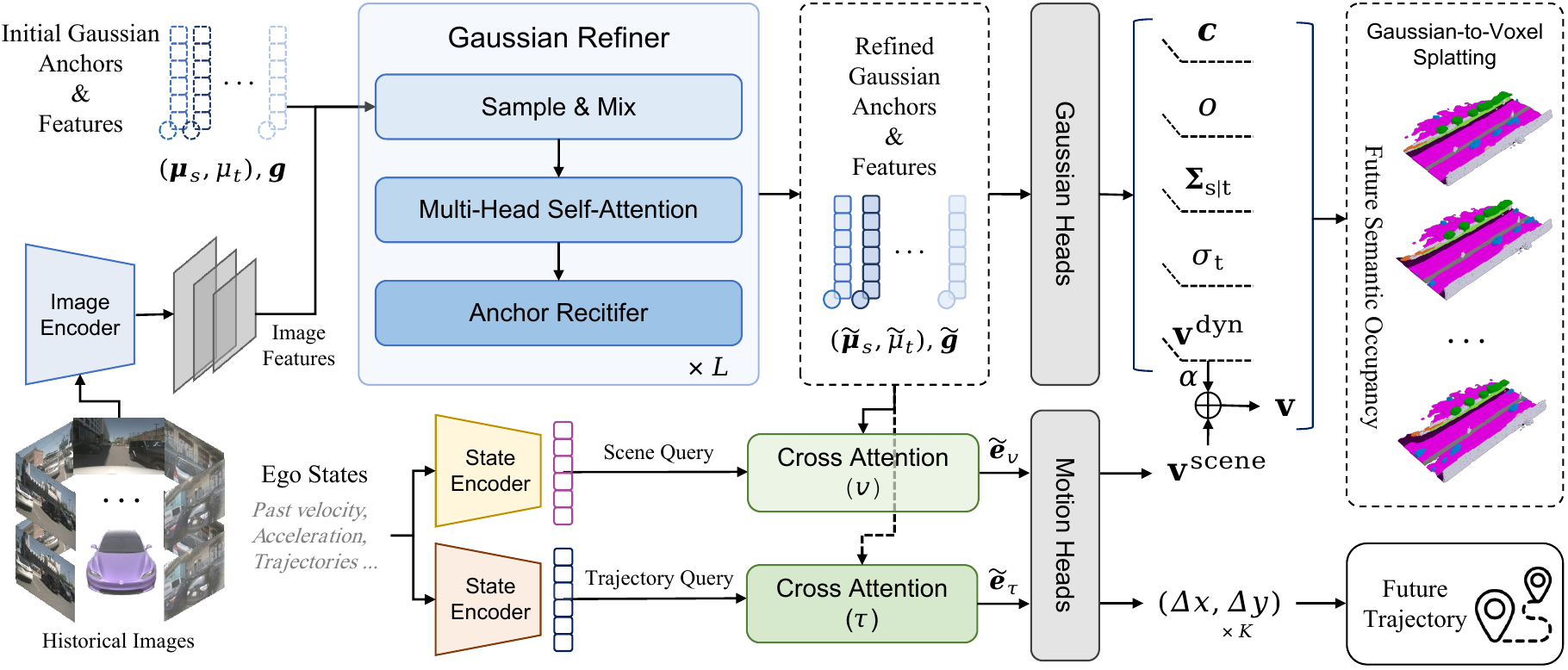}
    \caption{Overall pipeline of \name. Historical multi-view images and ego states are encoded to refine spatiotemporal Gaussian anchors and features. The refined Gaussians are decoded into semantic, geometric, temporal, and motion attributes, then sliced and splatted into future semantic occupancy volumes, while separate ego-conditioned queries support Gaussian velocity decomposition and future motion trajectory prediction.}
    \label{fig:pipeline}
\end{figure}
%

\subsection{\name: Occupancy Forecasting and Motion Planning}
\label{subsec:evoworld}

Figure~\ref{fig:pipeline} illustrates the overall \name pipeline. Each Gaussian primitive is represented by a spatiotemporal anchor and a latent feature. For the $q$-th primitive, we define:
\begin{equation}
\mathbf{a}_q = (\boldsymbol{\mu}_{s,q}, \mu_{t,q})\enspace, 
\qquad 
\mathbf{g}_q \in \mathbb{R}^{D}\enspace,
\label{eq:gaussian_anchor_feature}
\end{equation}
where $\boldsymbol{\mu}_{s,q}$ and $\mu_{t,q}$ denote its spatial center and temporal anchor, respectively. The anchor provides the geometric state for projection and refinement, while the feature $\mathbf{g}_q$ encodes the information used to predict Gaussian attributes, \ie, semantics, opacity, scale, rotation, motion, and temporal support. To model scene evolution in the moving ego frame, we further decompose each Gaussian velocity $\mathbf{v}_q$ defined in Equation~\eqref{eq:planar_velocity} into an ego-induced scene velocity $\mathbf{v}^{\mathrm{scene}}_q$ and an object-level dynamic velocity $\mathbf{v}^{\mathrm{dyn}}_q$. The former captures the apparent motion of the scene caused by ego vehicle movement, while the latter captures independent object motion.

\noindent\textbf{Gaussian Refiner.} To align Gaussian primitives with image observations before attribute decoding, we refine their anchors and features through stacked interaction blocks. Since the initial anchors provide only coarse spatiotemporal hypotheses, repeated interaction allows each primitive to gather visual evidence, exchange global context, and adjust its location before predicting semantic and geometric attributes. Each refinement block consists of image feature aggregation, efficient global interaction, and anchor rectification. 
First, each Gaussian primitive aggregates multi-view, multi-scale image features through adaptive spatiotemporal feature sampling and adaptive mixing~\cite{liu2023sparsebev}. Next, we use $M$ learnable latent features $\mathbf{f}_B\in\mathbb{R}^{M\times D}$, with $M\ll Q$, to model global interactions among the $Q$ Gaussian primitive features $\mathbf{f}_G=\{\mathbf{g}_q\}_{q=1}^{Q}\in\mathbb{R}^{Q\times D}$.
The multi-head attention is computed as:
\begin{equation}
\mathbf{f}_Z = \mathrm{MHAttn}(\mathbf{f}_B, \mathbf{f}_G, \mathbf{f}_G)\enspace,
\qquad
\tilde{\mathbf{f}}_G = \{\tilde{\mathbf{g}}_q\}_{q=1}^Q = \mathrm{MHAttn}(\mathbf{f}_G, \mathbf{f}_Z, \mathbf{f}_Z)\enspace.
\label{eq:induced_self_attention}
\end{equation}
Finally, each anchor is updated by a residual anchor rectifier:
\begin{equation}
(\Delta \boldsymbol{\mu}_{s,q}, \Delta \mu_{t,q})
=
\mathrm{MLP}_{\mathrm{rect}}(\tilde{\mathbf{g}}_q)\enspace,
\qquad
\tilde{\mathbf{a}}_q 
=
(\tilde{\boldsymbol{\mu}}_{s,q}, \tilde{\mu}_{t,q}) 
=
\mathbf{a}_q + (\Delta \boldsymbol{\mu}_{s,q}, \Delta \mu_{t,q})\enspace.
\label{eq:anchor_rectification}
\end{equation}
By repeating these blocks, \name progressively aligns Gaussian anchors and features with visual evidence while keeping geometric localization and attribute prediction decoupled.

\noindent\textbf{Gaussian Heads.}
Let $\tilde{\mathbf{g}}_q$ denote the final refined feature of the $q$-th Gaussian primitive. We decode it into the attributes needed for continuous-time slicing and occupancy splatting, including semantic logits, opacity, spatial scale, rotation, object-level dynamic velocity, and temporal support:
\begin{equation}
\begin{aligned}
&\mathbf{c}_q = \mathrm{MLP}_{c}(\tilde{\mathbf{g}}_q)\enspace,
\qquad
&o_q = \sigma\!\left(\mathrm{MLP}_{o}(\tilde{\mathbf{g}}_q)\right)\enspace,
\qquad
&\mathbf{s}_q = \mathrm{MLP}_{s}(\tilde{\mathbf{g}}_q)\enspace, \\
&\mathbf{r}_q = \mathrm{norm}\!\left(\mathrm{MLP}_{r}(\tilde{\mathbf{g}}_q)\right)\enspace,
\qquad
&\mathbf{v}^{\mathrm{dyn}}_q = \mathrm{MLP}_{v_d}(\tilde{\mathbf{g}}_q)\enspace,
\qquad
&\sigma_{t,q} = \mathrm{MLP}_{t}(\tilde{\mathbf{g}}_q)\enspace.
\end{aligned}
\label{eq:gaussian_heads}
\end{equation}
Here $\mathbf{s}_q$ and $\mathbf{r}_q$ parameterize the conditional spatial covariance $\mathbf{\Sigma}_{s\mid t,q}$, while $\sigma_{t,q}$ determines the primitive's temporal support. 

\noindent\textbf{Gaussian Velocity Decomposition.}
\label{sec:motion_decomposition}
Let $\mathbf{e}$ denote the ego state. As shown at the bottom of Figure~\ref{fig:pipeline}, we use a state encoder to project $\mathbf{e}$ into a scene query $\mathbf{q}_{v}=\psi_v(\mathbf{e})$, which attends to the refined Gaussian primitive features $\tilde{\mathbf{f}}_G$. The corresponding motion head $\mathrm{MLP}_{v}$ decodes the attended scene feature into the ego-induced scene velocity. Since the scene is represented in the ego frame, static scene elements have the opposite apparent velocity with respect to the ego vehicle:
\begin{equation}
\tilde{\mathbf{e}}_{v}
=
\mathrm{CrossAttn}_{v}
\left(
\mathbf{q}_{v},
\tilde{\mathbf{f}}_G,
\tilde{\mathbf{f}}_G
\right)\enspace,
\qquad
\mathbf{v}^{\mathrm{scene}}_q
=
-\mathrm{MLP}_{v}
\left(
\tilde{\mathbf{e}}_{v}
\right)\enspace.
\label{eq:scene_velocity}
\end{equation}
For primitives associated with dynamic semantic classes, we add the object-level dynamic velocity predicted by the Gaussian head in Equation~\ref{eq:gaussian_heads}, modulated by the primitive's dynamic probability:
\begin{equation}
\mathbf{v}_q
=
\mathbf{v}^{\mathrm{scene}}_q
+
\alpha_q \mathbf{v}^{\mathrm{dyn}}_q\enspace,
\label{eq:scene_dynamic_velocity}
\end{equation}
with $\alpha_q \in [0,1]$ estimated from the semantic distribution. Static primitives have small $\alpha_q$ and follow the ego-induced scene motion, while dynamic primitives have larger $\alpha_q$ and move independently.

\noindent\textbf{Motion Planning.}
To support safe motion planning, we use a separate state encoder to project the ego state into a trajectory query $\mathbf{q}_{\tau}=\psi_{\tau}(\mathbf{e})$, which attends to the refined Gaussian primitive features. The corresponding motion head $\mathrm{MLP}_{\tau}$ decodes the attended trajectory feature into future ego displacements. We stop gradients from the planning loss to the Gaussian features $\tilde{\mathbf{f}}_G$:
\begin{equation}
\tilde{\mathbf{e}}_{\tau}
=
\mathrm{CrossAttn}_{\tau}
\left(
\mathbf{q}_{\tau},
\mathrm{sg}(\tilde{\mathbf{f}}_G),
\mathrm{sg}(\tilde{\mathbf{f}}_G)
\right)\enspace,
\qquad
\hat{\boldsymbol{\tau}}_{1:K}
=
\mathrm{MLP}_{\tau}
\left(
\tilde{\mathbf{e}}_{\tau}
\right)\enspace.
\label{eq:traj_head}
\end{equation}
Here $\mathrm{sg}(\cdot)$ denotes stop-gradient, and
$\hat{\boldsymbol{\tau}}_{1:K}
=
\{\hat{\boldsymbol{\tau}}_k\}_{k=1}^{K}$ denotes the planned future ego trajectory, where each $\hat{\boldsymbol{\tau}}_k \in \mathbb{R}^{2}$ represents the incremental planar displacement $(\Delta x_k,\Delta y_k)$ at step $k$.

\noindent\textbf{Voxelization and Supervision.}
\label{sec:slicing_voxelization_supervision}
At a query timestamp $t$, each 4D Gaussian is sliced into a conditional 3D Gaussian, and the resulting primitives are splatted into a semantic occupancy volume using local aggregation, following GaussianFormer-2~\cite{gsformer2}. We supervise semantic occupancy with ground-truth semantics at each training timestamp using a combination of semantic cross-entropy and Lovasz loss~\cite{berman2018lovasz} for robust occupancy optimization:
\begin{equation}
\mathcal{L}_{\mathrm{occ}}
=
\lambda_{\mathrm{ce}}
\mathcal{L}_{\mathrm{ce}}
+
\lambda_{\mathrm{lov}}
\mathcal{L}_{\mathrm{lov}}\enspace.
\label{eq:occ_loss}
\end{equation}
For motion planning, we supervise the predicted future ego waypoints with ground-truth waypoints $\hat{\boldsymbol{\tau}}$ using mean squared error:
\begin{equation}
\mathcal{L}_{\mathrm{plan}}
=
\mathrm{MSE}
\left(
\hat{\boldsymbol{\tau}},
\boldsymbol{\tau}
\right)\enspace.
\label{eq:traj_loss}
\end{equation}
The final training objective is
$
\mathcal{L}
=
\mathcal{L}_{\mathrm{occ}}
+
\lambda_{\mathrm{plan}}
\mathcal{L}_{\mathrm{plan}}
$ where we set $\lambda_{\mathrm{plan}}$ to 1.0 in our experiments.
\section{Experiments}
\label{sec:exp}

\subsection{Dataset and Metrics}
We conduct experiments on the Occ3D-nuScenes benchmark~\cite{tian2023occ3d}, which provides dense semantic occupancy annotations for the nuScenes dataset~\cite{caesar2020nuscenes}. Each annotation covers a spatial range of $[-40, 40]\,\mathrm{m}$, $[-40, 40]\,\mathrm{m}$, and $[-1, 5.4]\,\mathrm{m}$ around the ego vehicle. The ground-truth occupancy is represented as a $200 \times 200 \times 16$ voxel grid with $0.4 \,\mathrm{m}$ resolution. Each voxel is labeled as one of 18 classes, including 17 semantic classes and one free-space class. Following common evaluation protocols, we report IoU and mIoU for 4D occupancy forecasting. IoU measures foreground-background occupancy overlap, while mIoU averages semantic IoU over the 17 semantic classes. For motion planning, we report L2 error and collision rate.

\subsection{Results and Analysis}

\noindent\textbf{4D Occupancy Forecasting.}
Table~\ref{tab:occ_res} compares \name with existing vision-centric occupancy world models for 4D occupancy forecasting. \name achieves the best average mIoU among vision-centric methods, improving over SparseWorld from $13.20$ to $13.60$. The gain is more clear in IoU, where \name obtains the best results at all future horizons and improves the average IoU from $22.03$ to $23.82$. We attribute the stronger IoU to \name's explicit continuous geometry: each Gaussian carries physical scene attributes and preserves occupied regions across future horizons through continuous spatiotemporal occupancy.

We also observe that SparseWorld~\cite{dang2026sparseworld} has slightly higher mIoU at the $3$s horizon, while \name achieves better average mIoU and consistently stronger IoU. One possible reason is that \name is trained end-to-end in a single stage, which requires the model to jointly balance near- and long-horizon supervision. Since near-future states are easier to infer from current observations, the optimization may favor shorter horizons, making fine-grained semantic prediction at $3$s more challenging.

\begin{table}[t]
\centering

\caption{\textbf{4D occupancy forecasting performance on the Occ3D-nuScenes dataset.} We use bold and underlined numbers to denote the best and second-best results, respectively.} 
\label{tab:occ_res}
\resizebox{\columnwidth}{!}{
\begin{tabular}{l c c cccc cccc}
\toprule
\multirow{2}{*}{Method} & \multirow{2}{*}{Input} & \multirow{2}{*}{Aux. Sup.} &
\multicolumn{4}{c}{mIoU $\uparrow$} &
\multicolumn{4}{c}{IoU $\uparrow$}\\
& & &
1s & 2s & 3s & Avg. &
1s & 2s & 3s & Avg.
\\
\midrule
\midrule
OccWorld-T~\cite{zheng2024occworld}        & Camera & LiDAR   & 4.68  & 3.36  & 2.63  & 3.56  & 9.32  & 8.23  & 7.47  & 8.34\\
OccWorld-D        & Camera           & 3D Occ   & 11.55 & 8.10  & 6.22  & 8.62  & 18.90 & 16.26 & 14.43 & 16.53\\
PreWorld~\cite{li2025semisupervised}  & Camera        & 3D Occ           & 11.69 & 8.72  & 6.77  & 9.06  & 23.01 & 20.79 & 18.84 & 20.88\\
\quad + Pre-training     &Camera & 2D \& 3D Occ   & 12.27 & 9.24  & 7.15  & 9.55  & \underline{23.62} & 21.76 & 19.63 & 21.62\\
SparseWorld~\cite{dang2026sparseworld} & Camera & 3D Occ         & \underline{14.93} & \underline{13.15} & \textbf{11.51} & \underline{13.20} & 22.96 & \underline{22.10} & \underline{21.05} & \underline{22.03} \\
\midrule
\name (Ours) & Camera& 3D Occ & \textbf{15.71} & \textbf{13.62} & \underline{11.46} & \textbf{13.60} & \textbf{25.56} & \textbf{24.03} & \textbf{21.86} & \textbf{23.82} \\
\midrule
\midrule
OccLLaMA-F~\cite{wei2024occllama}        & Camera+Action+Language & 3D Occ            & 10.34 & 8.66  & 6.98  & 8.66  & 25.81 & 23.19 & 19.97 & 22.99\\
GWM~\cite{GWM} & Camera+LiDAR & 3D Occ & \underline{11.63} & \underline{10.07} & \underline{8.17} & \underline{10.12} & \underline{26.22} & \underline{24.97} & \underline{22.13} & \underline{24.60}\\
\midrule
\name (Ours) & Camera, with LiDAR init & 3D Occ &  \textbf{18.66} & \textbf{15.93} & \textbf{13.02} & \textbf{15.87} & \textbf{31.56} & \textbf{28.94} & \textbf{25.42} & \textbf{28.64} \\
\bottomrule
\end{tabular}}
\end{table}

\begin{table}[!ht]
\centering
\caption{\textbf{Motion planning performance on the Occ3D-nuScenes dataset.} 
We use bold and underlined numbers to denote the best and second-best results, respectively.
}
\label{tab:planning_comparison}
\setlength{\tabcolsep}{10pt}
\resizebox{\columnwidth}{!}{%
\begin{tabular}{l c cccc cccc}
\toprule
\multirow{2}{*}{Method} & \multirow{2}{*}{Aux. Sup.} &
\multicolumn{4}{c}{L2 (m) $\downarrow$} &
\multicolumn{4}{c}{Collision Rate (\%) $\downarrow$} \\
& & 1s & 2s & 3s & Avg. & 1s & 2s & 3s & Avg. \\
\midrule
\midrule
VAD-Tiny~\cite{jiang2023vad} & Map \& Box \& Motion
& 0.46 & 0.76 & 1.12 & 0.78
& 0.21 & 0.35 & 0.58 & 0.38 \\

VAD-Base & Map \& Box \& Motion
& 0.41 & 0.70 & 1.05 & 0.72
& \textbf{0.07} & \underline{0.17} & \underline{0.41} & \underline{0.22}\\
\midrule
OccWorld-D~\cite{zheng2024occworld} & 3D Occ
& 0.52 & 1.27 & 2.41 & 1.40
& 0.12 & 0.40 & 2.08 & 0.87 \\
PreWorld~\cite{li2025semisupervised} & 3D Occ & 
0.22 & 0.31 & 0.41 & 0.31
& 0.36 & 0.52 & 0.73 & 0.54 \\

\quad + Pre-training & 2D \& 3D Occ
& 0.22 & 0.30 & 0.40 & 0.31
& 0.21 & 0.66 & 0.71 & 0.53 \\

SparseWorld~\cite{dang2026sparseworld} & 3D Occ
& \textbf{0.19} & \textbf{0.25} & \underline{0.36} & \textbf{0.27}
& 0.11 & 0.29 & 0.46 & 0.29 \\
\midrule
\name (Ours) & 3D Occ
& \textbf{0.19} & \underline{0.26} & \textbf{0.35} & \textbf{0.27}
& \underline{0.09} & \textbf{0.14} & \textbf{0.26} & \textbf{0.16} \\
\bottomrule
\end{tabular}%
}
\end{table}

\noindent\textbf{Motion Planning.}
Table~\ref{tab:planning_comparison} presents the motion planning results of \name. \name achieves an average L2 error of $0.27$m, matching the best camera-based result, and obtains the lowest average collision rate of $0.16\%$. Compared with SparseWorld, \name keeps similar trajectory accuracy while reducing the average collision rate from $0.29\%$ to $0.16\%$, with clear gains at $2$s and $3$s.

We attribute the lower collision rate to the explicit and interpretable Gaussian world representation. Since the trajectory head reasons over scene-level Gaussian primitives, the planner can use a structured representation of occupied space rather than only latent tokens or dense features. This helps the model produce safer trajectories by better modeling collision-relevant scene structure.

\noindent\textbf{Ablation Studies.}
Table~\ref{tab:ablation_num_gaussians} studies the effect of the number of Gaussian primitives. Reducing the number from $25{,}600$ to $12{,}800$ preserves similar forecasting performance, with slightly higher IoU and mIoU, while planning performance changes only marginally. This suggests that \name is not highly sensitive to moderate reductions in primitive count. However, further reducing the representation to $6{,}400$ Gaussians leads to a clear drop in occupancy forecasting accuracy and higher planning error, indicating insufficient scene coverage with too few primitives.

\begin{table}[h]
\centering
\begin{minipage}{0.47\linewidth}
\centering
\caption{Ablation on the number of Gaussians.}
\label{tab:ablation_num_gaussians}
\resizebox{\linewidth}{!}{
\begin{tabular}{ccccc}
\toprule
\multirow{2}{*}{\# Gaussians}
& \multicolumn{2}{c}{Forecasting}
& \multicolumn{2}{c}{Planning} \\
\cmidrule(lr){2-3} \cmidrule(lr){4-5}
& IoU $\uparrow$
& mIoU $\uparrow$
& L2 $\downarrow$
& Col. $\downarrow$ \\
\midrule
6400 & 20.91 & 12.12 & 0.29 & 0.24 \\
12800 & 23.99 & 13.62 & 0.28 & 0.18 \\
25600 & 23.82 & 13.60 & 0.27 & 0.16 \\
\bottomrule
\end{tabular}
}
\end{minipage}
\hfill
\begin{minipage}{0.47\linewidth}
\centering
\caption{Ablation on the core design choices. 
}
\label{tab:ablation_comp}
\resizebox{\linewidth}{!}{
\begin{tabular}{lcccc}
\toprule
\multirow{2}{*}{Method}
& \multicolumn{2}{c}{Forecasting}
& \multicolumn{2}{c}{Planning} \\
\cmidrule(lr){2-3} \cmidrule(lr){4-5}
& IoU $\uparrow$
& mIoU $\uparrow$
& L2 $\downarrow$
& Col. $\downarrow$ \\
\midrule
\name (full)
& 23.82 
& 13.60 
& 0.27 
& 0.16 \\
w/ unified vel.
& 23.23 
& 12.88 
& 0.37 
& 0.24 \\
w/ full 4D cov.
& 21.83 
& 11.42 
& 0.27 
& 0.20 \\
\bottomrule
\end{tabular}
}
\end{minipage}
\end{table}

We also examine the effect of incorporating modalities beyond camera images in the bottom section of Table~\ref{tab:occ_res}. LiDAR-based initialization significantly improves the IoU and mIoU of \name, showing that stronger geometric priors can improve Gaussian placement and lead to more accurate future occupancy forecasting. Compared with prior methods using extra inputs such as LiDAR, action, or language, \name achieves stronger performance, showing that the Gaussian representation can effectively use additional geometric cues.

Table~\ref{tab:ablation_comp} presents the ablation study evaluating the contribution of core designs. Replacing the decomposed motion model with a unified velocity field degrades forecasting performance and increases trajectory error, supporting the separation between ego-induced scene motion and object-level dynamic residual motion. In addition, a full 4D covariance formulation leads to worse occupancy performance, suggesting that the unconstrained spatiotemporal covariance is harder to optimize than our structured formulation.

\begin{figure}[!t]
    \centering
    \includegraphics[width=.94\columnwidth]{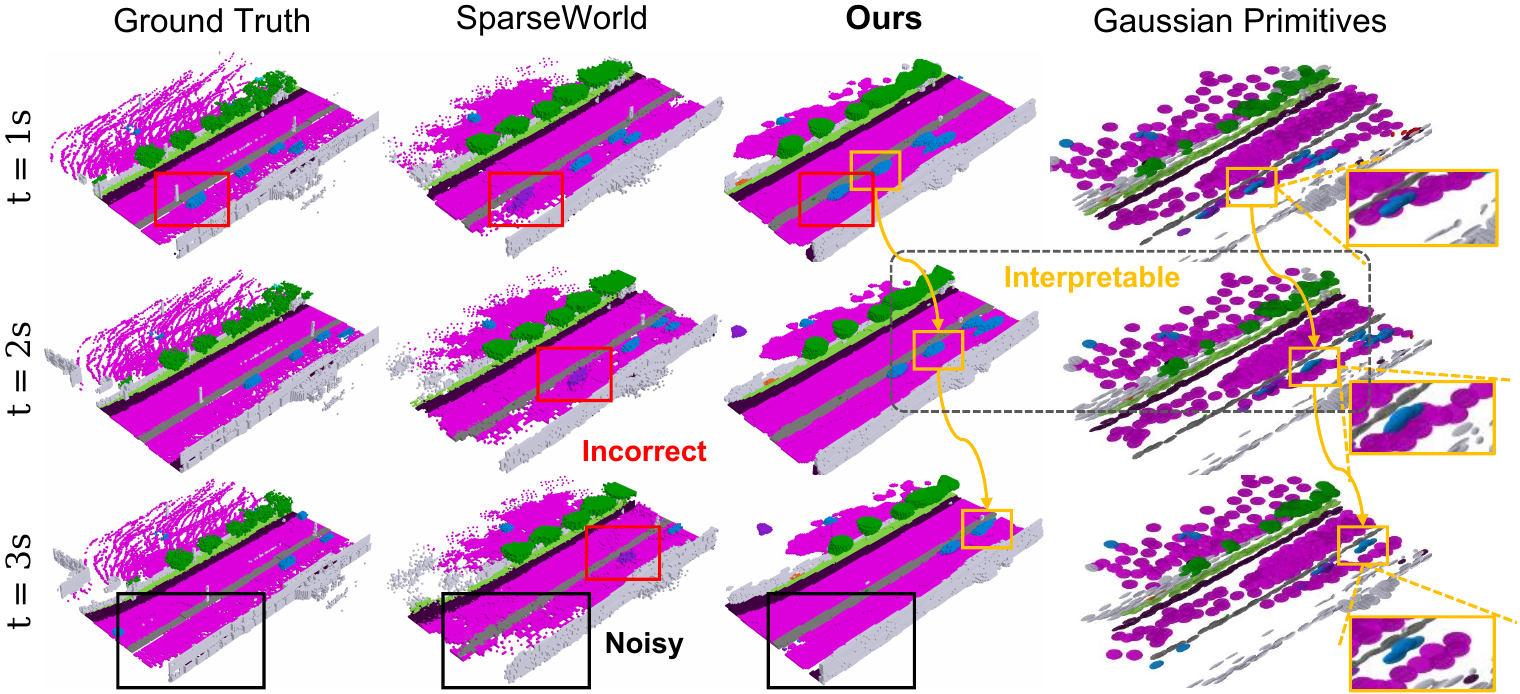}
    \caption{Qualitative comparison of future occupancy forecasting at $1$--$3$ seconds. Compared with SparseWorld~\cite{dang2026sparseworld}, \name produces cleaner and more accurate occupancy predictions over time. The right column shows the corresponding Gaussian primitives, where the highlighted blue car primitives remain identifiable, making the predicted motion directly inspectable.}
    \label{fig:vis_comparison}
\end{figure}

\noindent\textbf{Visualizations.}
Figure~\ref{fig:vis_comparison} compares future occupancy predictions from $1$s to $3$s. SparseWorld~\cite{dang2026sparseworld} captures the global layout but exhibits boundary artifacts and scattered occupied voxels near object surfaces, likely due to its point-query formulation. In contrast, \name's explicit continuous Gaussians provide smoother spatial support and more consistent temporal evolution, while keeping the underlying semantic and geometric primitives inspectable.

\begin{figure}[t]
    \centering
    \includegraphics[width=.94\columnwidth]{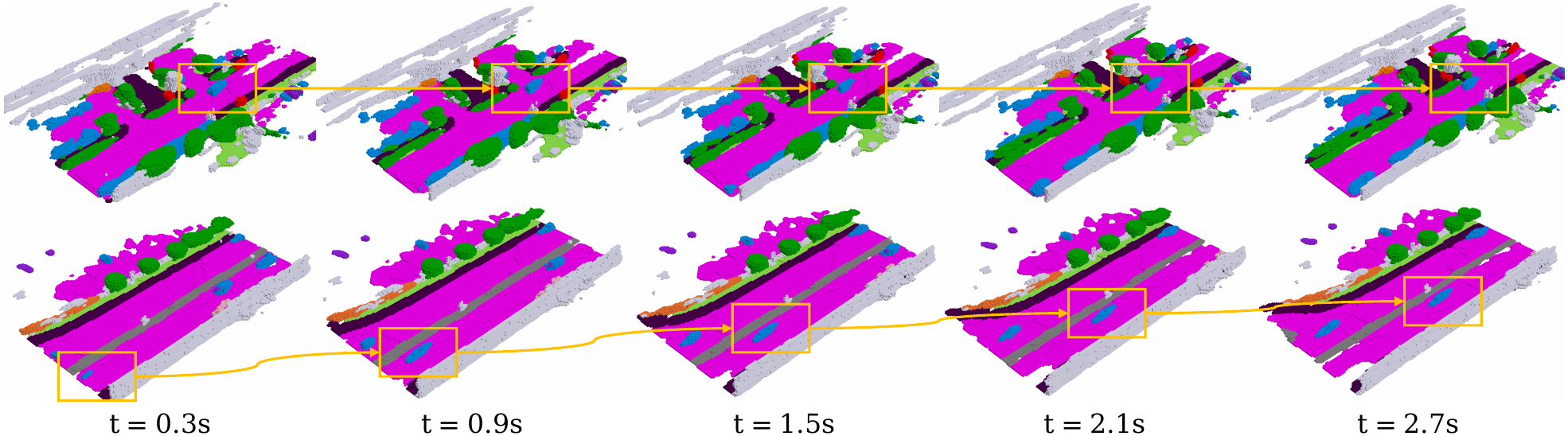}
    \caption{
    Arbitrary-timestamp occupancy forecasting of two scenes with \name. 
    The continuous 4D Gaussian world can be queried at any intermediate timestamp without sequential rollout, producing smooth scene evolution while preserving the global layout.
    }
    \label{fig:vis_arbitrary_query}
\end{figure}

Figure~\ref{fig:vis_arbitrary_query} illustrates the temporal flexibility of our continuous formulation. Unlike autoregressive methods with predefined $0.5$s output intervals, \name can be directly queried at arbitrary timestamps without sequential intermediate generation. The predictions evolve smoothly and preserve the global layout, enabling intermediate future occupancy outputs beyond fixed timestamps.
\section{Conclusion}

We present \name, a vision-centric, non-autoregressive occupancy world model based on continuous 4D Gaussian primitives. Instead of relying on autoregressive voxel or token rollout, \name directly queries a structured Gaussian scene representation at arbitrary future timestamps. This design enables dense semantic occupancy forecasting, interpretable scene evolution, and ego vehicle motion planning within a unified end-to-end framework. Extensive experiments demonstrate state-of-the-art 4D occupancy forecasting and strong motion planning performance, while the explicit Gaussian representation provides improved temporal flexibility and interpretability for dynamic driving scenes.

\bibliographystyle{unsrtnat}
\bibliography{references}


\appendix

\section{Related Work}
\label{sec:related}

\paragraph{3D occupancy prediction.}
3D occupancy prediction~\cite{tpvformer} has become an important scene representation for autonomous driving because it provides dense geometric and semantic understanding beyond object-level detection. In the vision-centric setting, a large body of work predicts semantic occupancy from multi-view cameras using BEV-, plane-, or volume-based representations~\cite{surroundocc,tpvformer,occformer,cvtocc,stcocc,alocc}. Along a different direction, Gaussian-based methods replace dense discretization with sparse scene primitives. The GaussianFormer series~\cite{huang2024gaussianformer,gsformer2} shows that semantic Gaussians can serve as compact occupancy primitives. GaussianWorld~\cite{zuo2025gaussianworld} extends this idea to streaming scene modeling, while GaussianOcc~\cite{gaussianocc} and GaussianFlowOcc~\cite{gaussianflowocc} show that Gaussian representations naturally support rendering- and projection-based occupancy learning. Recent work further extends Gaussian occupancy representations to collaborative perception~\cite{chen2026vision}. \name builds on this Gaussian line, but moves beyond 3D occupancy estimation toward a structured 4D Gaussian world representation for continuous future scene evolution.

\paragraph{Gaussian splatting for dynamic scene representation.}
Gaussian splatting has become a strong explicit scene representation for real-time novel view synthesis, starting from 3D Gaussian Splatting~\cite{3dgs} and extending to dynamic settings through time-varying positions, deformation fields, or 4D Gaussian parameterizations~\cite{dynamic3dgs,yang2023gs4d,4drotor,asiimwe20254d}. This line mainly targets view synthesis, video reconstruction, and dynamic scene rendering, where the goal is to reproduce photorealistic observations from different viewpoints and times. Recent urban-scene extensions such as Street Gaussians~\cite{streetgaussians} improve reconstruction quality in driving scenes, but still focus on rendering rather than semantic scene understanding or future occupancy prediction. \name builds on the compact, explicit, and continuous nature of Gaussian representations, but moves from rendering-oriented dynamic scene modeling to structured 4D semantic Gaussian occupancy world modeling for future scene forecasting.

\paragraph{Occupancy world models for autonomous driving.}
Occupancy world models extend 3D occupancy prediction from scene understanding to future scene evolution. Existing methods mainly follow two lines: predicting future occupancy from dense or sparse spatial representations, such as voxel features, BEV feature maps, and query-based states~\cite{driveoccworld,li2025semisupervised,dang2026sparseworld,dome,come}; or encoding occupancy into discrete latent tokens and predicting future tokens before decoding them back to occupancy volumes~\cite{zheng2024occworld,liao2025i2,wei2024occllama,occllm,yan2025renderworld}. Although effective, these methods commonly model future dynamics in latent spaces rather than explicit scene primitives, and predict future states at fixed steps autoregressively, which limits temporal flexibility and increases inference cost. A recent Gaussian-based world model introduces Gaussian priors into occupancy forecasting, but still predicts future scenes in a tokenized state space rather than using Gaussians as the world state itself~\cite{GWM}. In contrast, \name represents driving scenes directly as structured 4D semantic Gaussian primitives, which can be queried at arbitrary timestamps and splatted into conditional 3D Gaussians for explicit, continuous-time, and non-autoregressive occupancy forecasting.

\section{Implementation Details}

Following established protocols~\cite{zheng2024occworld,li2025semisupervised,dang2026sparseworld}, we use a 2-second historical context to forecast the next 3 seconds. We set the number of Gaussians to $Q=25{,}600$ and the feature channel dimension to $256$. The Gaussian refiner contains three interaction blocks, each using $M=1{,}280$ learnable latent features. We use ResNet-50 as the image encoder and a 3-layer MLP as the state encoder. We train \name end-to-end for 20 epochs with AdamW~\cite{adamw} and a cosine annealing learning-rate scheduler. The loss weights $\lambda_{\mathrm{ce}}$, $\lambda_{\mathrm{lov}}$, and $\lambda_{\mathrm{plan}}$ are set to $6.0$, $1.0$, and $1.0$, respectively. The initial learning rate is $2\times10^{-4}$, the weight decay is $0.01$, and the batch size is 1 per GPU on 2 H100 GPUs. During inference, \name can directly query any future timestamp within the 3-second horizon without autoregressive rollout.

\section{Limitations and Future Work}
While \name demonstrates strong performance with an explicit and continuously queryable 4D Gaussian world representation, several directions remain open. First, our current initialization study mainly considers two endpoints: random Gaussian initialization using only camera inputs, and LiDAR-based initialization using an additional geometric modality. A useful next step is to study camera-only initialization strategies between these endpoints, such as depth-guided placement or priors from foundation models, which may improve Gaussian coverage without relying on LiDAR. Second, \name currently predicts occupancy over a fixed set of semantic classes. Extending the 4D Gaussian world model with language grounding could enable open-world forecasting, richer scene understanding, and more flexible planning.

\section{Broader Impacts}

\name targets vision-centric occupancy forecasting and motion planning for autonomous driving. By producing explicit and continuously queryable scene representations, it may improve the transparency and temporal flexibility of driving perception systems. These properties could support safer planning, better failure analysis, and more interpretable debugging in autonomous vehicles. In the long term, such methods may contribute to more capable autonomous driving systems. However, deployment should require extensive closed-loop testing, uncertainty estimation, and human oversight. This work uses public driving datasets and does not introduce new privacy or surveillance risks beyond those already associated with such datasets.

\end{document}